\documentclass[journal]{vgtc}                     

\onlineid{0}

\vgtccategory{Research}

\vgtcpapertype{please specify}

\title{SceneLinker: Compositional 3D Scene Generation via \\ Semantic Scene Graph from RGB Sequences}

\author{%
  \authororcid{Seok-Young Kim}{0009-0008-7336-5699},
  \authororcid{Dooyoung Kim}{0000-0002-6003-2181},
  \authororcid{Woojin Cho}{0009-0003-4615-5630},
  \authororcid{Hail Song}{0009-0006-4008-196X},
  \authororcid{Suji Kang}{0009-0001-1614-6245}, and  
  \authororcid{Woontack Woo}{0000-0002-5501-4421}
}

\authorfooter{
  \item
  	Seok-Young Kim is with KAIST UVR Lab.
  	E-mail: seokyoung@kaist.ac.kr.
  \item
  	Dooyoung Kim is with KAIST KI-ITC ARRC.
  	E-mail: dooyoung.kim@kaist.ac.kr.
  \item
  	Woojin Cho is with KAIST KI-ITC PMRC.
  	E-mail: woojin.cho@kaist.ac.kr.
  \item
  	Hail Song is with KAIST UVR Lab.
  	E-mail: hail96@kaist.ac.kr.
  \item
  	Suji Kang is with KAIST UVR Lab.
  	E-mail: forzestlife@kaist.ac.kr.
  \item Woontack Woo is with KAIST UVR Lab and KAIST KI-ITC ARRC.
  	E-mail: wwoo@kaist.ac.kr.
}

\abstract{%
  We introduce SceneLinker, a novel framework that generates compositional 3D scenes via semantic scene graph from RGB sequences. 
To adaptively experience Mixed Reality (MR) content based on each user's space, it is essential to generate a 3D scene that reflects the real-world layout by compactly capturing the semantic cues of the surroundings.
Prior works struggled to fully capture the contextual relationship between objects or mainly focused on synthesizing diverse shapes, making it challenging to generate 3D scenes aligned with object arrangements.
We address these challenges by designing a graph network with cross-check feature attention for scene graph prediction and constructing a graph-variational autoencoder (graph-VAE), which consists of a joint shape and layout block for 3D scene generation. 
Experiments on the 3RScan/3DSSG and SG-FRONT datasets demonstrate that our approach outperforms state-of-the-art methods in both quantitative and qualitative evaluations, even in complex indoor environments and under challenging scene graph constraints. 
Our work enables users to generate consistent 3D spaces from their physical environments via scene graphs, allowing them to create spatial MR content. 
Project page is \url{https://scenelinker2026.github.io}.
}


\teaser{
  \centering
  \includegraphics[width=\linewidth]{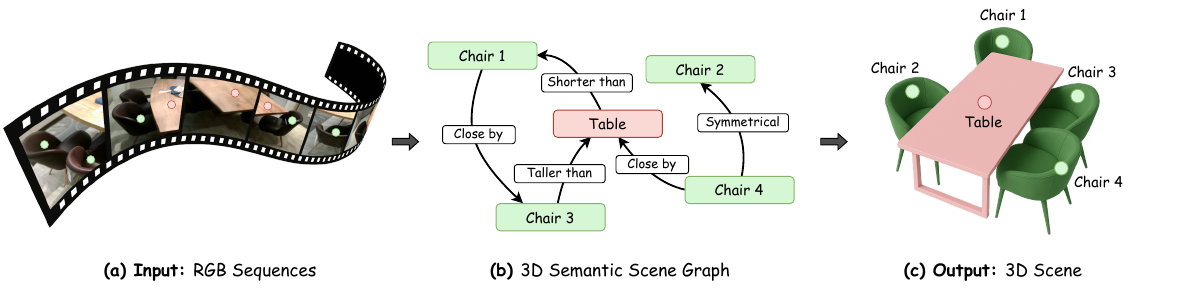}
  \caption{%
  	\textbf{SceneLinker} generates compositional 3D scenes by concisely representing real spaces using a graph structure. Our approach takes (a) RGB sequences as input, incrementally predicts a (b) global 3D scene graph, and constructs a (c) 3D scene by generating 3D objects corresponding to the scene graph using a graph-VAE.%
  }
  \label{fig:teaser}
}

\graphicspath{{figs/}{figures/}{pictures/}{images/}{./}} 

\usepackage{tabu}                      
\usepackage{booktabs}                  
\usepackage{lipsum}                    
\usepackage{mwe}                       

\usepackage{mathptmx}                  
\usepackage{times}                     

\usepackage{amssymb}
\usepackage{multirow}
\usepackage{amsmath}
\usepackage{enumitem}
\usepackage{pifont}
\usepackage{makecell} 
\usepackage{placeins} 
\usepackage[table,xcdraw]{xcolor}
\usepackage{xcolor} 

\begin{document}


\firstsection{Introduction}

\maketitle

With the recent release of technologies such as Meta's SceneScript \cite{scenescript} and Apple's RoomPlan\footnote{https://developer.apple.com/augmented-reality/roomplan/}, which scan and virtualize the real world, the demand for immersive 3D applications utilizing space as a medium is growing in Augmented Reality (AR) and Virtual Reality (VR) \cite{intro2, intro3, intro5}. To provide a realistic experience by reflecting the user's space in Mixed Reality (MR) content, it is essential to leverage the geometry and semantic information of the real-world, enabling more natural and intuitive interactions.
One of the key technologies to enable this is the virtualization of physical environments, which involves reconstructing a 3D asset of the real world \cite{m2,rw2, realitycrafter} or generating virtual 3D scenes while preserving the actual spatial layout for scalability \cite{scenescript, vipscene}.
In addition, to fully leverage real-world details in MR content, it is crucial to have an efficient data structure capable of managing and storing scene information while preserving spatial characteristics. The scene graph, a 3D scene understanding representation that combines the semantic and relational information of objects, has proven its effectiveness in managing spatial information \cite{3dsg,plan_sg1, sg_his}. This compact structure is well-suited for compressing complex 3D information into text and providing an intuitive visual representation that users can easily understand and interact with \cite{sg_ar, sgbot, sg-tailor, graphdreamer}. 
In addition, to achieve seamless MR authoring content, there is emerging attention to connecting real and virtual spaces via scene graphs and generating compositional virtual scenes tailored to each user's environment \cite{meta_obj, design, imaginatear}.

Many studies have applied scene graphs for linking user's physical environment. However, two key challenges remain in using them to represent spaces in MR contents. First, the object-level graph has not been fully utilized to connect real and virtual scenes. While object-level relationships can enable compositional layout and efficient instance management, prior approaches mainly rely on room-scale relationships \cite{kimera, hydra, khronos}. In robotics, scene graphs have been used to represent hierarchical-level 3D environments using open-vocabulary \cite{hier_extra, sayplan, con_sg}, but these methods fail to capture pairwise object relationships, limiting their use in interactive AR/VR settings. In addition, a graph-based efficient framework for generating 3D scenes from RGB sequences has rarely been addressed in previous works, making it challenging to create compositional 3D scenes that reflect user-specific spatial arrangements. Second, existing graph-based 3D generative methods struggled to arrange a coherent layout among virtual objects, largely due to their prioritization of generating various object shapes \cite{common, echo, mmg}. 
Although the positional information of the layout plays a dominant role in generating a complete scene that corresponds to the scene graph, previous methods have not sufficiently incorporated this into the training scheme. 
Consequently, they failed to generate 3D scenes that correspond to the graph’s semantic relationships.

To address these challenges, we propose SceneLinker, a comprehensive framework that takes RGB sequences as input and generates a layout-consistent 3D scene via a pairwise scene graph.
Our approach consists of two phases: 3D scene graph prediction and 3D virtual scene generation. The first phase incrementally constructs segmented maps and point clouds using Visual-SLAM \cite{orbslam3} from image sequences to estimate 3D entities. These entities form nodes in a neighbor graph represented by oriented bounding boxes, multi-view images, and point features. We then apply an attention-based graph network to propagate robust message features between nodes and refine probability weights, resulting in a global scene graph. In the second phase, the constructed graph and predicted bounding boxes are embedded into our scene generation network. Inspired by \cite{common}, we enhance object and relationship features using a pre-trained vision-language model \cite{clip}, and assign initial shape priors to nodes using DeepSDF \cite{deepsdf}. Our Graph-VAE with a Joint Shape and Layout block (JSL block) fuses correlated shape and bounding box features, stably preserving object positions in graph-based generated scenes.

We evaluate the proposed method on the real 3D scene datasets 3RScan/3DSSG \cite{3rscan,3DSSG} and the SG-FRONT dataset \cite{common}, which include object-level mesh data \cite{3d_front} annotated with semantic scene graphs. Our approach demonstrates superior performance compared to existing methods in both scene graph prediction and 3D scene generation quality. Notably, in the scene graph consistency experiments, our model achieved performance improvements of over 7\% and 14\% on challenging relationship metrics such as ``\texttt{close by}'' and ``\texttt{symmetrical}'' respectively. 
This graph-level approach enables the capturing of surrounding entities and their spatial relationships within an AR/VR environment, while seamlessly generating virtual content linked to the real-world layout.

Our main contributions can be summarized as follows:
\begin{itemize}
    \item We propose SceneLinker, a novel method for generating 3D virtual scenes from RGB sequences by predicting a 3D scene graph, enabling consistent and manageable scene representation. 

    \item We design a Cross-Check Feature Attention (CCFA) network tailored for 3D scene graph prediction, which enhances robustness in complex environments by adaptively refining attention weights through node-wise message passing.

    \item We construct a 3D scene generation network built on a Joint Shape and Layout block (JSL block), which improves scene-level consistency and outperforms existing state-of-the-art methods.

\end{itemize}

\section{Related Work}
\subsection{Image-based 3D Generative Methods}
Recently, the task of generating high-quality 3D assets from images has emerged as a key focus in AR/VR and computer vision communities.
Leveraging a 2D diffusion prior, \cite{rw1, rw2} introduced an initial baseline method for producing 3D assets from a single image.
Subsequently, to enhance single‑view object consistency, methods \cite{m1, m2, m5, m7} have been proposed that synthesize diffusion‑based multi‑view images and then reconstruct a single 3D object. Following on these advancements, \cite{dit1, dit2, dit3} apply diffusion transformer \cite{dit} to generate 3D geometry with generalization across diverse datasets through extensive training. However, previous works mainly focused on reconstructing a single object from an image, suffering to generate multiple objects and capture precise compositional relationships between instances. To address this limitation, \cite{reparo, midi, sam3d} advanced the field by using segmentation \cite{sam} to reconstruct compositional multiple objects from a single image. Nevertheless, these methods were not robust to occlusions of object back views and could only generate fully visible objects from a single image frame. 
Moreover, prior reconstruction-focused methods struggled to generate diverse shapes and to achieve the fast inference speed critical for MR environments.
Our work leverages a graph-based representation to robustly capture multi-object features and their compositional relationships by taking a sequence of multi-view images as input. Additionally, we aim to enable users to generate diverse 3D scenes with minimal latency by retrieving shapes or synthesizing objects based on the real-world layout.

\subsection{3D Scene Graph Prediction}
In the field of AR/VR, understanding 3D scenes with minimal resources and effectively managing semantic information for real-time interaction is essential. To this end, scene graphs have become a key tool in efficient computer vision tasks, serving as a compact and explicit representation for scene understanding. Specifically, in the study of 3D scene graphs applicable to the head-mounted display systems, the method introduced in \cite{vgfm} serves as a representative baseline, improving upon the 2D image-based method discussed in \cite{imp}. By incorporating geometric information extracted from multiple views and iteratively passing a tripartite graph, it estimates visual graphs from sequential images. However, this method is computed based on a limited view of the images, leading to inconsistencies in object relationships when aggregated into a global graph. To address this issue, \cite{sgfn} proposed a method that utilizes geometric segmentation based on semantic SLAM \cite{denseslam} with RGB-D sequences, enabling real-time incremental 3D scene graph prediction. With these advances, \cite{MONOSSG} utilizes sparse points extracted from ORB-SLAM3 \cite{orbslam3} based on RGB images to generate a scene graph. By fusing image and point features for message passing, it demonstrates the effectiveness of combining these two complementary modalities. However, while this previous work has focused on extracting robust 3D entity features, it has not fully addressed the propagation of semantic messages within the scene graph.

To address this, \cite{twin} proposed a twinning attention mechanism that applies attention to both nodes and edges to enhance their interactions. Building on this approach, \cite{twin++} proposed a scene graph-aware attention module that combines twinning attention with cross-modal attention to improve communication between language and graph modalities. However, this module was specifically designed for only multi-modal tasks. Additionally, \cite{sgfn} introduced feature-wise attention, which proved beneficial for handling partial and incomplete 3D data as well as dynamic edges. Nonetheless, it had limitations in fully reflecting the relationships between neighboring nodes and edge features, as the attention value was focused primarily on the node features.
In this work, we design a robust scene graph prediction network for complex environments using two complementary modalities and a feature aggregation mechanism that thoroughly checks the features between neighboring nodes. Similar to \cite{sgfn}, our graph network applies attention on a feature-wise basis, but unlike \cite{sgfn}, we overcome existing limitations by cross-checking information from adjacent nodes that are connected by an edge to compute similarity scores.

\begin{figure*} [ht]
  \includegraphics[width=\textwidth]{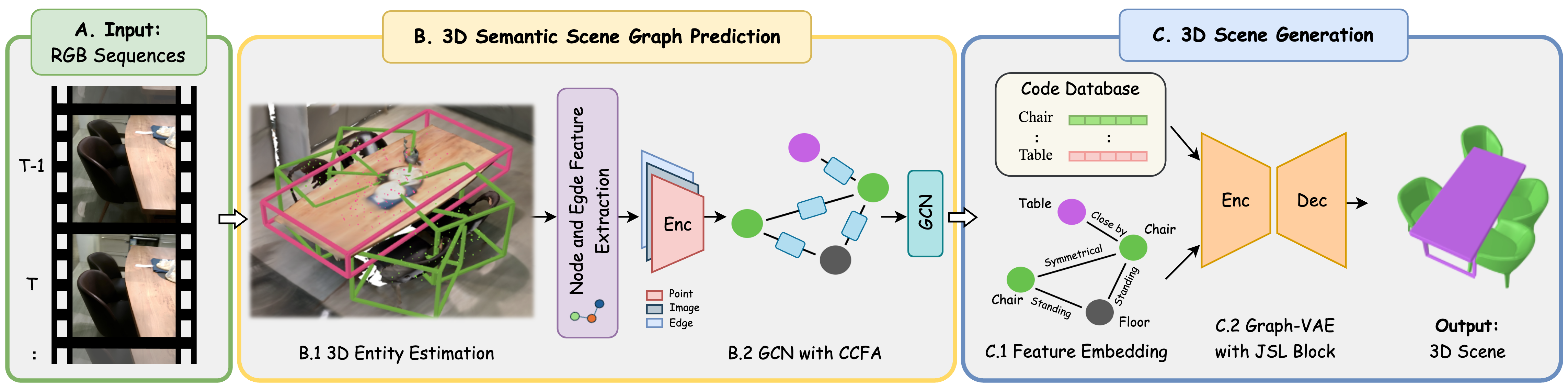}
  \caption{
    Given (A) RGB sequences as input to SceneLinker, the system estimates incremental (B.1) 3D entities. It then computes the node and edge features for each 3D segment to extract graph properties and propagates them to predict the (B.2) global scene graph. The (C.1) predicted scene graph and SDF shape code are embedded into a (C.2) GCN-based VAE, which fuses the shape and layout to finally generate a 3D scene.
  }
  \label{fig:overview}
\end{figure*}

\subsection{3D Scene Generation from Scene Graph}
In AR applications, generating virtual objects that correspond to real-world entities is a significant challenge for facilitating interaction between users and objects. Approaches that use scene graphs to generate 3D models have been extensively researched to manage 3D objects within the user's space in a resource-efficient manner \cite{g23d,common,echo,3dsln,instructscene}. By including an oriented bounding box in each node of the scene graph, which contains the size, position, and rotation of objects, this approach allows the generation of 3D scenes based on a compact graph structure. \cite{3dsln} utilizes a scene graph with oriented bounding box information and synthesizes a layout using differentiable rendering. However, because it does not consider the object's shape, it relies on shape database retrieval to generate a complete scene. This retrieval process can be resource-intensive in terms of computational time and memory usage, depending on the size of the database. Therefore, recent generative models have been proposed to simultaneously generate layout and shape \cite{g23d,common,echo, mmg}.

\cite{g23d} was the first to propose an end-to-end VAE-based network for synthesizing scenes and layouts jointly. It learns pre-trained shapes and layouts through separate encoders, then extracts a latent vector using a shared network, enabling scene manipulation based on a GAN. \cite{common} preprocesses the scene graph using CLIP \cite{clip} to construct a contextual graph, and then aggregates the features of each node with a graph-convolutional network (GCN). It also adjusts the conditions for shape generation through a diffusion-based denoising process. While \cite{common} relies on VAE and GAN for scene manipulation, similar to \cite{g23d}, both models struggle with synchronized learning for layout generation. 
To address this issue, \cite{echo} introduces a dual-branch diffusion model that generates layout and shape simultaneously. By employing an information echo method specialized for dynamic scene graphs in diffusion-based layout generation, it achieved high-quality results in both shape generation and scene manipulation. 
\cite{mmg} introduces a mixed-modality graph that combines image data with scene graphs and takes advantage of the dual-branch diffusion model from \cite{echo} to synthesize geometrically controllable 3D scenes.

These diffusion-based methods \cite{common, echo, mmg} excel at scene manipulation and shape diversity, but fail to generate scenes that consistently align with the scene graph input. 
Furthermore, the inherently slow inference speed of diffusion steps renders them unsuitable for time-sensitive MR applications.
Unlike prior works, SceneLinker addresses the limitations of scene graph consistency by merging shape and layout with a focus on bounding box information. We also adopt a VAE structure that enables relatively fast inference, thereby minimizing latency compared to existing diffusion models.

\section{Methodology}
In this section, we provide details of SceneLinker, the method for predicting scene graphs and generating 3D scenes. 
Our goal is to represent the real world at the graph-level and to generate context-consistent 3D scenes.
\cref{fig:overview} shows the overall framework diagram of the proposed SceneLinker. Our pipeline consists of two separate processes: the scene graph prediction network with cross-check attention (\cref{sec:3.1}), and layout-centric 3D scene generation (\cref{sec:3.2}).

\subsection{3D Semantic Scene Graph Prediction}
\label{sec:3.1}
In this process, we incrementally represent scenes as graphs by fusing a local graph for the current frame with a global graph updated from previous frames, enabling accurate recognition of multi-object environments. Our process is divided into an incremental 3D entity estimation module and an attention-based GCN module. The former stage estimates 3D entities based on image data and extracts semantic properties of nodes and edges. In the latter stage, the graph properties are used to predict the relationships between objects using a novel attention mechanism to complete the global scene graph.

\subsubsection{Incremental 3D Entity Estimation}
\label{sec:3.1.1}

To construct a global scene graph from a sequence of RGB images, it is necessary to incrementally perceive the scene and estimate the entities within the space. We adopt the incremental entity estimation method from \cite{MONOSSG}, which matches keypoints from sequential images based on ORB-SLAM3 \cite{orbslam3} to extract segmented sparse points. These points are then integrated into 3D entities, constructing an entity visibility graph and a neighbor graph. The entity visibility graph $G_{vis}(n,f,e)$ is a bipartite graph with nodes representing entities $n$, keyframes $f$, and visibility edges $e$. A visibility edge is generated when a node is visible in a keyframe. Additionally, the neighbor graph $G_{adj}(n,e)$ is an undirected graph consisting of nodes and their potential neighboring edge features, where the neighborhood is determined by the collision detection between oriented bounding boxes. 
These two graphs serve complementary roles in the subsequent graph reasoning process.
The entity visibility graph provides multi-view visual feature aggregation across observed keyframes, while the neighbor graph defines the spatial adjacency for message passing.

Each node with entity properties includes a multi-view image feature $R \in \mathbb{R}^2$ and a point feature $P \in \mathbb{R}^3$. 
The $r_i \in R$ is extracted by aggregating information from keyframes provided by the entity visibility graph, and encoded via a pretrained ResNet18 \cite{resnet18}.
Since the set of $p_i \in P$ is sparse and incomplete, it is processed through the PointNet \cite{pointnet} encoder. The refined $p_i \in \mathbb{R}^2$ are then integrated with the image features using a sigmoid-based learnable gate function $\rho$, to fuse the two complementary feature sets. To extract edge features, the point clouds of the two adjacent segments are first embedded by a PointNet \cite{pointnet}, and the relationship between them is subsequently determined by estimating the geometric relative pose of the nodes using the pose descriptor \cite{MONOSSG}. This process calculates the relative maximum and minimum values along each axis based on the corner coordinates of the bounding boxes. The computed features are then passed through a Multi-Layer Perceptron (MLP).
The node and edge features for message passing in the graph network are as follows:

\begin{equation}
N_i = (r_i+\rho(p_i), bbox^d_i, bbox^c_i, bbox^\alpha_i)
\label{eq:node}
\end{equation}

\begin{equation}
E_{ij} = MLP([N_i, N_j], D(bbox_i, bbox_j))
\label{eq:edge}
\end{equation}
In \cref{eq:node}, node feature ${N}_i$ includes features of the image and gated point, along with the property information of the oriented bounding box $bbox$: dimension $d$, centroid $c$, and angle value $\alpha$. The edge feature $E_{ij}$ in \cref{eq:edge} is computed based on the pose descriptor $D$, which encodes the oriented bounding box information of the two connected nodes $N_i$ and $N_j$.

\subsubsection{GCN with Cross-Check Feature Attention (CCFA)}

In this part, we predict the scene graph based on the nodes and edges extracted through entity estimation in \cref{sec:3.1.1}. Since the performance of scene graph prediction depends heavily on how the GCN propagates messages, it is crucial to design the network to aggregate and update the properties of nodes and edges effectively. 
To achieve this, we adopted an attention mechanism, a design strategy well-suited for maintaining consistency between the existing local graph and newly incoming information while incrementally constructing the global graph.
Inspired by \cite{sgfn}, our method applies attention based on node features, unlike \cite{sgfn}, we focus on aggregating messages considering the attention weights of both the source and target nodes sharing an edge. Additionally, we designed a scene graph-specific network to extract reliable features by sharing weights across neighboring nodes. 

Our CCFA-based GCN takes the initial graph features as input and maps the source node $N_i$, target node $N_j$, and their connecting edge $E_{ij}$ into a unified embedding space using a MLP. This step aligns node and edge feature dimensions to facilitate attention computation. The attention weights are then computed using the features of both connected nodes ($N_i$, $N_j$) and the corresponding edge ($E_{ij}$). Formally, the attention weight $W_{ij}$ between a pair of nodes $N_i$ and $N_j$, connected by edge $E_{ij}$, is computed as follows:

\begin{equation}
l(Q, K, V) = \text{softmax}\left(\frac{QK^T}{\sqrt{d_k}}\right) V
\label{eq:att}
\end{equation}

\begin{equation}
W_{ij} = l(N_i, E_{ij}, N_j) + l(N_j, E_{ij}, N_i)
\label{eq:CCFA1}
\end{equation}
$l$ denotes the attention function, which computes similarity scores based on node-edge features from query $Q$, key $K$, value $V$ and $d_k$ is dimension of key. After that, the softmax function normalizes these attention scores into probabilities. Thus, the attention mechanism explicitly integrates both nodes' perspectives via shared edge features, effectively achieving a ``cross-check" effect.
Once attention weights are obtained, the enhanced message representation \(M_{ij}\) for the target node $N_j$ is computed by Hadamard product between the node features $N_j$ and the attention weights $W_{ij}$:

\begin{equation}
M_{ij} = W_{ij} \odot N_j
\label{eq:CCFA2}
\end{equation}

In this manner, CCFA adaptively integrates relationships between neighboring nodes and focuses on reliable feature interactions, making it robust to incomplete or noisy inputs. Moreover, we employ multi-head attention as introduced in \cite{attention}, allowing parallel computation of multiple attention distributions to enhance feature representation capability.
Finally, at the last layer of the graph network, we integrate Gated Recurrent Units (GRU) \cite{gru} to iteratively update node and edge features according to graph topology. This iterative refinement ensures robust scene graph prediction by adaptively aggregating messages and reinforcing semantic reliability, even under challenging conditions involving multiple objects or incomplete data.

\subsubsection{Scene Graph Fusion}

Since the nodes and edges of the incrementally constructed partial local graph are predicted multiple times, we use a strategy similar to \cite{sgfn, MONOSSG} to globally integrate them. We update the same entities and predicates across temporal graphs using a moving average approach. 
At each time step $t$, the scene graph prediction network outputs a class probability $o^t$ for each node and edge. The probability $o^t$ is obtained by normalizing the predicted class logits using a softmax function.
To ensure temporal consistency, all attributes in the graph store a probabilistic weight $\varphi$. When a feature with probability $u$ is predicted at time $t$, it is fused based on the following equation: 

\begin{equation}
    u^t = \frac{u^t \cdot \varphi^t + u^{t-1} \cdot \varphi^{t-1}}{\varphi^t + \varphi^{t-1}}
    \label{eq:fusion1}
\end{equation}

\begin{equation}
    \varphi^t = \min \left( \varphi_{\text{max}}, \varphi^t + \varphi^{t-1} \right)
    \label{eq:fusion2}
\end{equation}
$\varphi_{max}$ is the maximum weight value, and set to 100, consistent with previous works \cite{sgfn, MONOSSG}. Finally, the predicted entity segments are integrated into a global scene graph, ensuring consistency across the entire 3D scene by utilizing these strategies.

\begin{figure}
\centering
 \includegraphics[width=1.0\columnwidth]{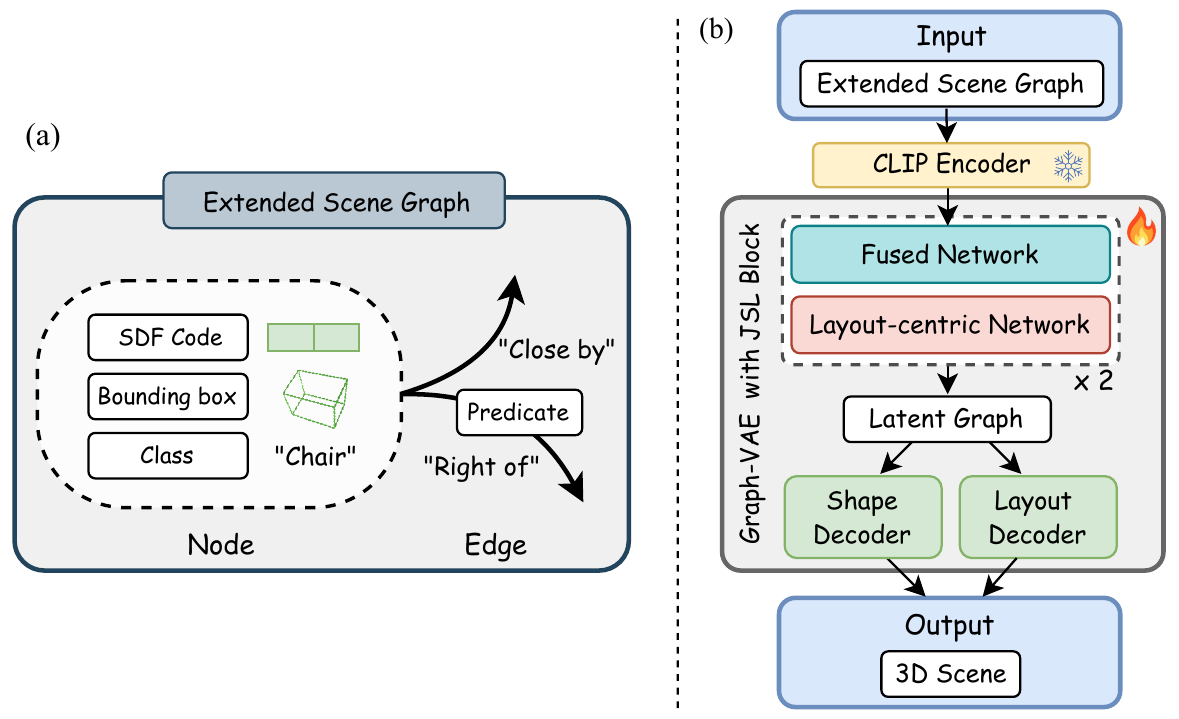}
 \caption{
 Configuration of (a) extended scene graph and proposed (b) graph-VAE sturcture for encoding, decoding. 
 }
 \label{fig:VAE with JSL}
\end{figure}

\subsection{Graph-based Compositional 3D Scene Generation}
\label{sec:3.2}

This section describes the method to generate a 3D scene based on a predicted scene graph. 
The approach aims to identify and generate the user’s surroundings while preserving the layout.
We first embed an initial scene graph with object properties and then compute the latent graph for both layout and shape using our JSL block-based encoder. Finally, the sampled latent is fed into the decoder to generate the complete 3D scene.

\subsubsection{Extended 3D Scene Graph Construction}
\label{sec:3.2.1}

Constructing a 3D scene from a graph requires each node’s object class, shape, and bounding box features. To generate a wider variety of virtual objects from existing scene compositions, we adopt a strategy that leverages prior shape information. Therefore, we expand the graph by incorporating an initial shape code for each corresponding node using an SDF-based shape generator as shown in \cref{fig:VAE with JSL}(a). Specifically, DeepSDF \cite{deepsdf} is employed for both encoding and decoding shapes.
This implicit function-based algorithm learns a continuous shape space, allowing the scene to be represented as a mesh.
Furthermore, inspired by \cite{common}, we enhance contextual features (e.g. class, predicate) in the graph using the pre-trained vision-language model, CLIP \cite{clip}. By enriching textual context, this approach captures object and relationship features more robustly than a conventional scene graph.
Our extended graph $G_{extend}(\hat{N}_i,\ \hat{E}_{ij},\ \hat{N}_j)$ is represented as follows:

\begin{equation}
    {\nu}_i = E_{CLIP}(obj^{cls}_{i}), \hspace{0.6em}
    {\nu}_{ij} = E_{CLIP}(obj^{cls}_{i},E_{ij},obj^{cls}_{j})
    \label{eq:vlm}
\end{equation}

\begin{equation}
\begin{gathered}
    G_{\text{extend}}(
    \hat{N}_i,\ \hat{E}_{ij},\ \hat{N}_j), \\
    \hat{N}_i = [obj^{\text{cls}}_{i} \oplus \nu_i,\ obj^{b}_{i},\ obj^{s}_{i}], \quad
    \hat{{E}_{ij}} = [E_{ij} \oplus \nu_{ij}]
\end{gathered}
\label{eq:extended graph}
\end{equation}

$E_{CLIP}$ is a pretrained text encoder, ${\nu}_i$ and ${\nu}_{ij}$ refer to the CLIP features of the node and edge, respectively. Each extended node $\hat{N_i}$ includes an object class $obj^{cls}_{i}$ concatenated with the CLIP feature ${\nu}_i$, an oriented bounding box $obj^{b}_{i}$, and latent shape code $obj^{s}_{i}$. The edge feature $\hat{E_{ij}}$ augmented with the contextual feature ${\nu}_{ij}$ identifies spatial relationships between objects.

\subsubsection{Graph-VAE with Joint Shape and Layout Block (JSL block)}
\label{sec:3.2.2}

Our encoder backbone is based on DeepGCN \cite{deepgcn} and employs two strategies for handling the extended scene graphs. First, since the shape and bounding box features of an object corresponding to each node are related and closely connected, we implement a training scheme that effectively fuses these two modalities. Secondly, given that the object shape is filled based on the bounding box during the 3D scene generation, we consider that layout information holds a more significant role than shape information. Accordingly, our JSL block is designed with a fused network to integrate these features and a layout-centric network that dominantly focuses on learning the bounding box, as shown in \cref{fig:VAE with JSL}(b).

The fused network of the JSL block aims to combine the initial shape and layout modalities completely. To achieve this, the embedded shape and layout vectors of the object and predicate are added element-wise, and fed into the 5 layers of GCN. By combining these two related features within the graph network, the modalities can interact and convey information. The layout-centric network aims to learn the features of the scene by focusing more on the layout of the bounding box. This module takes the coarse feature output from the fused network and the corresponding bounding box information as input again. It then refines the features focused on the bounding box through 5 layers of GCN. We combine the two modalities into a single-branch structure, employing a simple yet powerful network design that results in coarse-to-fine processing.

We designed the model to enable more robust encoding by using the proposed JSL block as the basic unit and adding a skip connection between two blocks. In the final step of the JSL block, an MLP computes the mean and standard deviation, incorporating the angle of the bounding box. These values are then used to form a normal distribution from which the latent graph $z$ is sampled.
To generate a 3D scene based on the latent graph, we decode each shape and layout in parallel using the strategy from \cite{g23d}. The layout decoder takes $z$ as input, passes it through a GCN, and then is split into two separate MLPs to predict the location and angle of the bounding box, respectively. The shape decoder follows a similar process but uses a single MLP to predict shapes. Subsequently, the 3D object shape is generated using the pre-trained DeepSDF shape generator. Then, each shape is transformed into global scene coordinates based on the predicted bounding boxes to complete the 3D scene.

\subsubsection{Training Loss}
\label{sec:3.2.3}

To train our proposed 3D scene generation model, the total loss comprises two primary components: the reconstruction loss and the Kullback-Leibler (KL) divergence loss, as defined in \cref{eq:loss_recon} and \cref{eq:loss_kl}, respectively. To reconstruct the 3D shape, we minimize $L_{recon}$ by comparing the predicted shape $s$, bounding box $bbox$, and angle $\alpha$ with the ground truth (GT) for supervision. $CEE$ is the Cross Entropy Error, which is used to classify the rotation angle.

\begin{equation}
     L_{recon} = \frac{1}{N} \sum_{i=1}^{N} \left( ||s_i - \hat{s_i}||_1 +  + ||bbox_i - \hat{bbox_i}||_1 + CEE(\alpha_i, \hat{\alpha}_i) \right)
    \label{eq:loss_recon}
\end{equation}

\begin{equation}
    L_{KL} = D_{KL} \left( {E}(z|x, bbox, s) \| p(z|x) \right)   
    \label{eq:loss_kl}
\end{equation}

$L_{KL}$ measures the difference between the probability distribution generated by the encoder network $E$ and the prior distribution $p(z|x)$. This quantifies the divergence between the two distributions, and we minimize this loss to ensure that the model-generated distribution approximates the prior distribution closely. The loss weights $\lambda_1$ and $\lambda_2$ are set to 1.0 and 0.1, respectively. The final loss term is the sum of \cref{eq:loss_recon} and \cref{eq:loss_kl} and is formulated as follows:

\begin{equation}
    L  =  \lambda_1 L_{recon}  +  \lambda_2 L_{KL}
    \label{eq:loss_total}
\end{equation}

\section{Experiments}

\subsection{Settings}
\textbf{Datasets. }
We evaluate our scene graph prediction on the 3RScan /3DSSG \cite{3rscan,3DSSG} dataset, which includes 1,482 real-world RGB-D scans of 478 indoor environments with scene graph annotations. In addition, graph-based 3D scene generation is validated on the SG-FRONT dataset, which extends the 3D-FRONT dataset \cite{3d_front} by adding scene graph annotations to its high-quality indoor home environment meshes. It consists of 45K object instances and 15 relationships across three types of scenes. Note that 3RScan contains low-quality incomplete object meshes, making it difficult to evaluate 3D scene generation. Due to this issue, \cite{g23d} trained an object-level mesh using ShapeNet \cite{shapenet}. However, their method failed to generalize properly due to inconsistencies between the datasets. To ensure a clearer and more equitable comparison, we conduct 3D scene generation experiments using the SG-FRONT, as in \cite{common, echo, mmg}.

~\\ 
\textbf{Evaluation Metrics. }
To evaluate the accuracy of the predicted scene graph, we use the metrics outlined in \cite{MONOSSG}, including overall recall, which is commonly used in scene graph tasks, and mean recall to address class imbalance. Object and predicate metrics are calculated based on their respective classification scores, while the relationship score is determined by multiplying the probabilities of the object, predicate, and subject. For benchmarking, we compare our method's performance with existing approaches using the top-1 metric as the standard. Additionally, we assess 3D scene generation accuracy by following scene graph constraint metrics, as described in \cite{g23d,common,echo,mmg}. This evaluates how the generated scene aligns with the input scene graph based on layout relationships. Following prior works, we divide relations into easy metrics (\texttt{left/right}, \texttt{front/behind}, \texttt{smaller/larger}, \texttt{taller/shorter}) determined only by bounding box coordinates and hard metrics (\texttt{close by}, \texttt{symmetrical}) that require pairwise comparisons with explicit thresholds (0.45 as specified by SG-FRONT) for proximity or symmetry.

~\\
\textbf{Baselines. }
Our experiments are conducted using state-of-the-art methods according to benchmark of each module. For 3D scene graph estimation, we compare our methods with previous works as baselines \cite{imp,vgfm,3DSSG,sgfn,MONOSSG}. IMP \cite{imp}, VGFM \cite{vgfm}, MonoSSG \cite{MONOSSG} and our model are based on RGB sequences input, while 3DSSG \cite{3DSSG} uses point clouds and SGFN \cite{sgfn} utilizes RGB-D frames to predict 3D scene graphs. 
For graph-driven 3D scene generation methods, we adopt Graph-to-3D \cite{g23d}, which uses a GCN-based VAE backbone; CommonScenes \cite{common}, a dual-branch architecture using a VAE with an LDM \cite{ldm}; and EchoScene \cite{echo}, a dual-branch diffusion model that synchronizes layout and shape. We also compare with MMGDreamer \cite{mmg}, which leverages a mixed-modality graph that fuses scene graphs with object images.

~\\ 
\textbf{Implementation Details. }
The proposed model was implemented using the PyTorch framework and trained/evaluated on an Intel Core i9-10900 3.7GHz CPU with 20 threads and a single NVIDIA RTX 3090ti GPU (24GB VRAM). 
In the encoder for scene generation, class categories are fed into the embedding layer before GCN computation, and shape embeddings, boxes, and angles are projected through the linear layer.
We used the Adam optimizer for training the graph-VAE network, with an initial learning rate set to 0.001. 

\begin{table}[t]
\caption{
Scene graph prediction on 3RScan dataset with 20 object and eight predicate classes (higher is better). The table highlights the {\colorbox[HTML]{769CF6}{best}} and {\colorbox[HTML]{C5D2F8}{second}} performances.
}
\centering


\begin{tabular}{l|ccc|cc}
\hline
\multicolumn{1}{c|}{}                         & \multicolumn{3}{c|}{\textit{Recall(\%)}}                                                   & \multicolumn{2}{c}{\textit{mRecall(\%)}}                    \\
\multicolumn{1}{c|}{\multirow{-2}{*}{Method}} & Rel.                         & Obj.                         & Pred.                        & Obj.                         & Pred.                        \\ \hline
IMP \cite{imp}                                         & 46.2& 66.1& 94.1& 55.1& 45.6\\
VGFM \cite{vgfm}                                          & 52.0& 69.7& 94.4& 61.0& 48.7\\
3DSSG \cite{3DSSG}                                        & 30.3& 53.6& 95.2                         & 49.5& 58.5\\
SGFN \cite{sgfn}                                         & 42.6& 64.1& 94.2& 56.2& 63.7\\
MonoSSG \cite{MONOSSG}                                      & \cellcolor[HTML]{C5D2F8}63.7& \cellcolor[HTML]{C5D2F8}79.3& \cellcolor[HTML]{C5D2F8}95.6& \cellcolor[HTML]{C5D2F8}78.0& \cellcolor[HTML]{C5D2F8}64.5\\ \hline
Ours                                          & \cellcolor[HTML]{769CF6}\textbf{68.3}& \cellcolor[HTML]{769CF6}\textbf{81.4}& \cellcolor[HTML]{769CF6}\textbf{96.1}& \cellcolor[HTML]{769CF6}\textbf{79.6}& \cellcolor[HTML]{769CF6}\textbf{69.1} \\ \hline
\end{tabular}%

\label{tab:sg_20}
\end{table}

\begin{table}[]
\caption{Scene graph prediction on 3RScan dataset with 160 object and 26 predicate classes (higher is better). The coloring is consistent with \cref{tab:sg_20}.} 
\centering
\begin{tabular}{l|ccc|cc}
\hline
\multicolumn{1}{c|}{}                         & \multicolumn{3}{c|}{\textit{Recall(\%)}}                                                   & \multicolumn{2}{c}{\textit{mRecall(\%)}}                    \\
\multicolumn{1}{c|}{\multirow{-2}{*}{Method}} & Rel.                         & Obj.                         & Pred.                        & Obj.                         & Pred.                        \\ \hline
IMP \cite{imp}                                         & 61.5& 42.6& 13.1& 17.7& 4.5\\
VGFM \cite{vgfm}                                          & 63.3& 45.9& 14.6& 17.4& 6.1\\
3DSSG \cite{3DSSG}                                        & 61.7& 29.6& \cellcolor[HTML]{769CF6}\textbf{67.3}& 11.2& 24.1\\
SGFN \cite{sgfn}                                         & 64.1& 36.3& 48.5& 15.9& 14.0\\
MonoSSG \cite{MONOSSG}                                      & \cellcolor[HTML]{C5D2F8}64.9& \cellcolor[HTML]{C5D2F8}54.5& 48.4& \cellcolor[HTML]{C5D2F8}28.8& \cellcolor[HTML]{C5D2F8}24.5\\ \hline
Ours                                          & \cellcolor[HTML]{769CF6}\textbf{68.7}& \cellcolor[HTML]{769CF6}\textbf{55.6}& \cellcolor[HTML]{C5D2F8}53.9& \cellcolor[HTML]{769CF6}\textbf{29.7}& \cellcolor[HTML]{769CF6}\textbf{27.8}\\ \hline
\end{tabular}%

\label{tab:sg_160}
\end{table}

\begin{table*}[t]
\caption{Scene graph constraints on the \textbf{retrieval-based} methods (higher is better). To minimize class imbalance bias, the total accuracy is computed by averaging the accuracy across each edge class.}
\small
\centering
\resizebox{\textwidth}{!}{%
\begin{tabular}{lccccc|cc}
\toprule
\multicolumn{1}{c}{\multirow{2}{*}{Method}} &
\multirow{2}{*}{\begin{tabular}[c]{@{}c@{}}Shape\\ Representation\end{tabular}} &
\multicolumn{4}{c|}{Easy} & 
\multicolumn{2}{c}{\textbf{Hard\textasteriskcentered}} \\
\cmidrule(lr){3-6} \cmidrule(lr){7-8}
& & Left/right & Front/behind & Smaller/larger & Taller/shorter & \textbf{Close by\textasteriskcentered} & \textbf{Symmetrical\textasteriskcentered} \\
\midrule
3D-SLN \cite{3dsln}              &  & 0.97          & 0.99          & 0.95          & 0.91          & 0.72          & 0.47          \\
Progressive \cite{g23d}          &  & 0.97          & 0.99          & 0.95          & 0.82          & 0.69          & 0.46          \\
Graph-to-Box \cite{g23d}         & Retrieval  & 0.98          & 0.99          & 0.96          & 0.95          & 0.72          & 0.45          \\
CommonLayout \cite{common}       & in Database & 0.98          & 0.99          & \textbf{0.97} & 0.95          & 0.74          & 0.63          \\
EchoLayout \cite{echo}           &  & \textbf{1.00} & 0.99          & 0.95          & \textbf{0.96} & 0.74          & \textbf{0.67} \\
Ours                             &  & 0.98          & 0.99          & \textbf{0.97} & \textbf{0.96} & \textbf{0.75} & 0.58          \\
\bottomrule
\end{tabular}%
}
\label{tab:generation_1}
\end{table*}

\begin{table*}[t]
\caption{Scene graph constraints on the \textbf{generative-based} methods (higher is better). To minimize class imbalance bias, total accuracy is computed by averaging the accuracy across each edge class.}
\small
\centering
\resizebox{\textwidth}{!}{%
\begin{tabular}{lccccc|cc}
\toprule
\multicolumn{1}{c}{\multirow{2}{*}{Method}} &
\multirow{2}{*}{\begin{tabular}[c]{@{}c@{}}Graph\\ Modality\end{tabular}} &
\multicolumn{4}{c|}{Easy} & 
\multicolumn{2}{c}{\textbf{Hard\textasteriskcentered}} \\
\cmidrule(lr){3-6} \cmidrule(lr){7-8}
& & Left/right & Front/behind & Smaller/larger & Taller/shorter & \textbf{Close by\textasteriskcentered} & \textbf{Symmetrical\textasteriskcentered} \\
\midrule
Graph-to-3D \cite{g23d}  & Text& 0.98 & 0.99 & \textbf{0.97} & 0.95 & 0.74 & 0.57 \\
CommonScenes \cite{common} & Text& 0.98 & \textbf{1.00} & \textbf{0.97} & 0.95 & 0.77 & 0.60 \\
EchoScene \cite{echo}    & Text& 0.98 & 0.99 & 0.96          & 0.96 & 0.74 & 0.55 \\
MMGDreamer \cite{mmg}    & Text, Image& 0.98 & 0.99 & \textbf{0.97}          & 0.96 & 0.76 & 0.62 \\
Ours                     & Text& 0.98 & 0.99 & \textbf{0.97} & \textbf{0.97} & \textbf{0.82} & \textbf{0.71}\\

\bottomrule
\end{tabular}%
}

\label{tab:generation_2}
\end{table*}

\subsection{3D Semantic Scene Graph Estimation}

Following \cite{MONOSSG}, we conduct two quantitative evaluations by dividing 3RScan into a small version with 20 objects and eight predicate classes, and a large version with 160 objects and 26 predicate classes. For overall recall, we evaluate graph components from various perspectives: triplet relationship (Rel.) prediction composed of $subject-predicate-object$ sets, object (Obj.) prediction for nodes, and predicate (Pred.) prediction for edges. In the experiment with 20 objects and eight predicate classes, our model achieved superior performance, as reported in ~\cref{tab:sg_20}. Specifically, our model demonstrated performance metrics of 68.3\%, 81.4\%, and 96.1\% in Rel, Obj, and Pred recall, and achieved 79.6\% and 69.1\% accuracy in Obj and Pred mRecall. These results outperform state-of-the-art methods in all metrics. It suggests that our graph network propagates and updates enhanced messages more effectively than previous scene graph-specific attention techniques.

In addition, SceneLinker maintains robust performance in more complicated conditions. In experiments with 160 objects and 26 predicate classes, our model demonstrated the most stable accuracy overall, as shown in \cref{tab:sg_160}. We recorded recall scores of 68.7\%, 55.6\%, and 53.9\% for Rel, Obj, and Pred, respectively, ranking first for Rel and Obj, second for Pred in the benchmark. In this table, 3DSSG demonstrates outstanding performance in predicate prediction for overall recall, indicating that its point-based approach is specialized for predicate prediction. Notably, our model achieves mRecall scores of 29.7\% and 27.8\%, representing the highest performance. This result proves that our model is relatively reliable even when the input dataset has severe data imbalance issues. Thus, our network design enables iterative cross-checking of messages, ensuring reliable updates to node and edge properties, which leads to the construction of a robust scene graph for indoor entities.

\subsection{Graph-based 3D Scene Generation}

To evaluate the 3D scenes consistent with the scene graph input, we divide our experiments into two categories: retrieval approaches and methods that consider layout and shape together. Retrieval-based evaluation focuses on network performance with respect to layout only, as shapes are retrieved from the database based on the bounding box. As a baseline, 3D-SLN \cite{3dsln} generates a layout for the scene, while Progressive adds objects one by one sequentially based on \cite{g23d}. CommonLayout and EchoLayout also use object retrieval to construct scenes in \cite{common}, \cite{echo}. For shape generative-based methods, Graph-to-3D employs DeepSDF similar to our approach, whereas CommonScenes, EchoScene, and MMGDreamer use a truncated SDF \cite{TSDF} based on VQ-VAE \cite{vq-vae}.

In retrieval-based evaluation, our model achieves comparable performance with accuracies of 0.98, 0.99, 0.97, and 0.96 for the easy part, as shown in \cref{tab:generation_1}. Also, our model effectively captures close relationships with an accuracy of 0.75 for the hard part. However, for the symmetrical metric, the dual-branch diffusion method \cite{echo} outperformed our approach. In contrast, when considering both layout and shape, as shown in \cref{tab:generation_2}, our method demonstrates higher accuracy compared to approaches focusing only on layout. This suggests that our model, which integrates both shape and layout through a unified network, provides more consistent scene graph-based modeling when both features are incorporated during training. Our model achieved accuracies of 0.98, 0.99, 0.97, and 0.97 for the easy part, respectively. For the hard part, specifically the ``\texttt{close by}'' and ``\texttt{symmetrical}'' metrics, we improve accuracy by over 7\% and 14\% compared to the state-of-the-art model.    
These results validate the effectiveness of the JSL block’s coarse-to-fine process and confirm the benefits of integrating shape with a layout-centric focus.

\begin{table}[]
\caption{Ablation study of the JSL block under three conditions.}
\centering
\scalebox{1.0}{%
    \begin{tabular}{@{}cccccc@{}}
    \toprule
\multirow{2}{*}{Model} & \multirow{2}{*}{CLIP} & \multirow{2}{*}{\begin{tabular}[c]{@{}c@{}}$Fused_N$\end{tabular}} & \multirow{2}{*}{\begin{tabular}[c]{@{}c@{}}$Layout_N$\end{tabular}} & \multicolumn{2}{c}{\textbf{Hard\textasteriskcentered}} \\ \cmidrule(l){5-6}
                       &                      &                      &                      & \textbf{Close by\textasteriskcentered}  & \textbf{Symmetrical\textasteriskcentered}  \\ \midrule
(a)                    & \ding{55}            & \ding{55}            & \checkmark          & 0.752         & 0.573            \\
(b)                    & \ding{55}           & \checkmark           & \ding{55}           & 0.763         & 0.632            \\
(c)                    & \ding{55}            & \checkmark           & \checkmark          & 0.791         & 0.674\\
(d)                    & \checkmark           & \checkmark           & \checkmark          & \textbf{0.824} & \textbf{0.708}\\ \bottomrule
    \end{tabular}%
}
\label{tab:ablation}
\end{table}

\begin{figure*}[!ht]
\centering
\includegraphics[width=\linewidth]{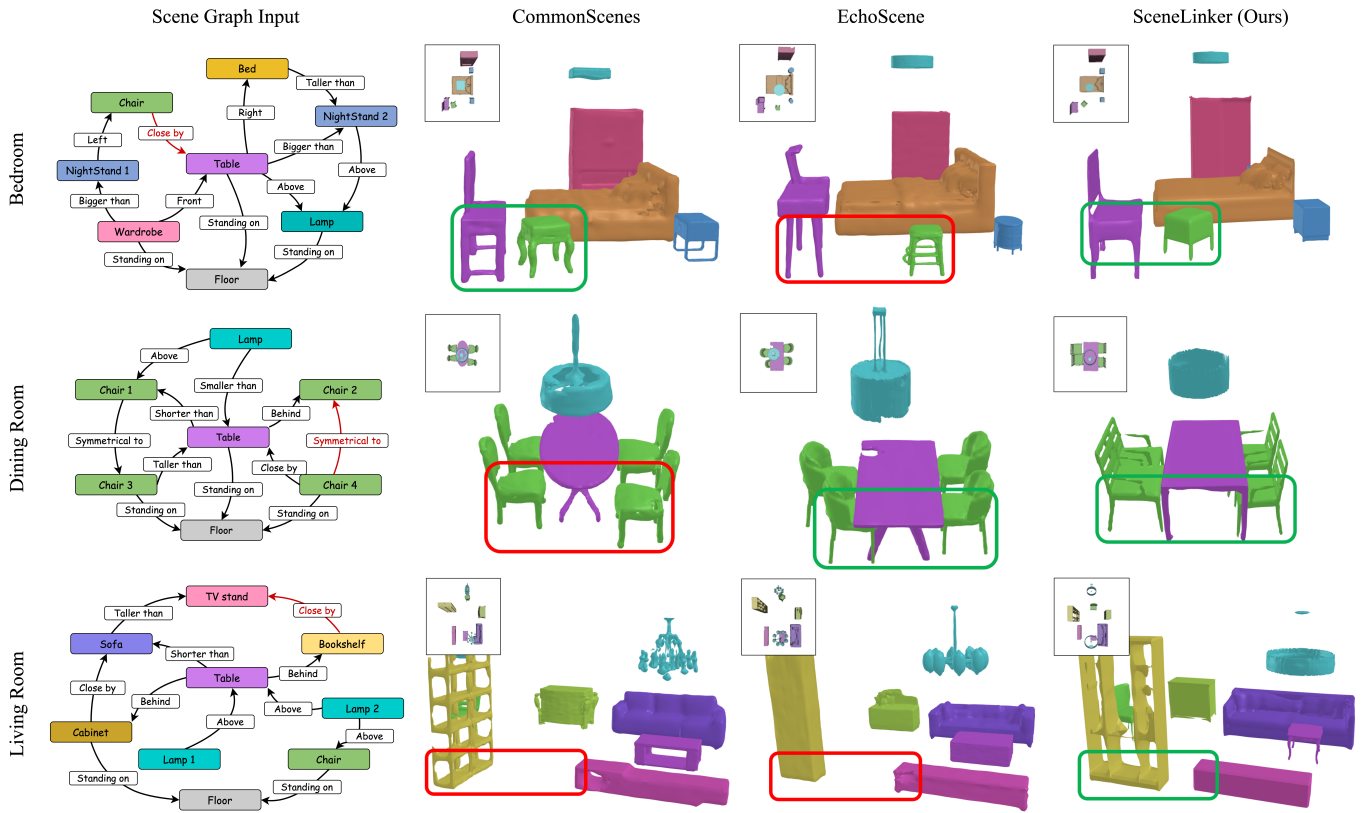}
\caption{
    Qualitative comparisons with other generative models.  Our method is compared with approaches that utilize text-based graph. For the challenging scene graph constraints (\texttt{close by}, \texttt{symmetrical}), correctly generated areas are highlighted in green, while incorrectly generated areas are highlighted in red.
}
\label{fig:qualitative}
\end{figure*}

To analyze the impact of the JSL block, we conduct an ablation study in three parts as shown in \cref{tab:ablation}. 
Since our fused network receives both shape and layout as input, for study (a), we encode both features in parallel in separate GCNs. As a result, the model in (a), which learns the two features separately, shows the lowest performance. This demonstrates that employing a joint feature-sharing network is reliable, as shape and layout are closely interconnected. In study (b), where we exclude the layout-centric refinement network, the model achieved scores of 0.763 and 0.632. Although this shows improvement over the model (a), it does not reach the performance of our full JSL block model (c), which achieved 0.791 and 0.674. Finally, embedding graphs with CLIP preprocessing along with JSL blocks achieved the best accuracy with 0.824 and 0.708. This result suggests that enhanced contextual features and bounding boxes have a more significant impact on 3D scene generation.

\begin{figure*}[!ht]
\centering
\includegraphics[width=\linewidth]{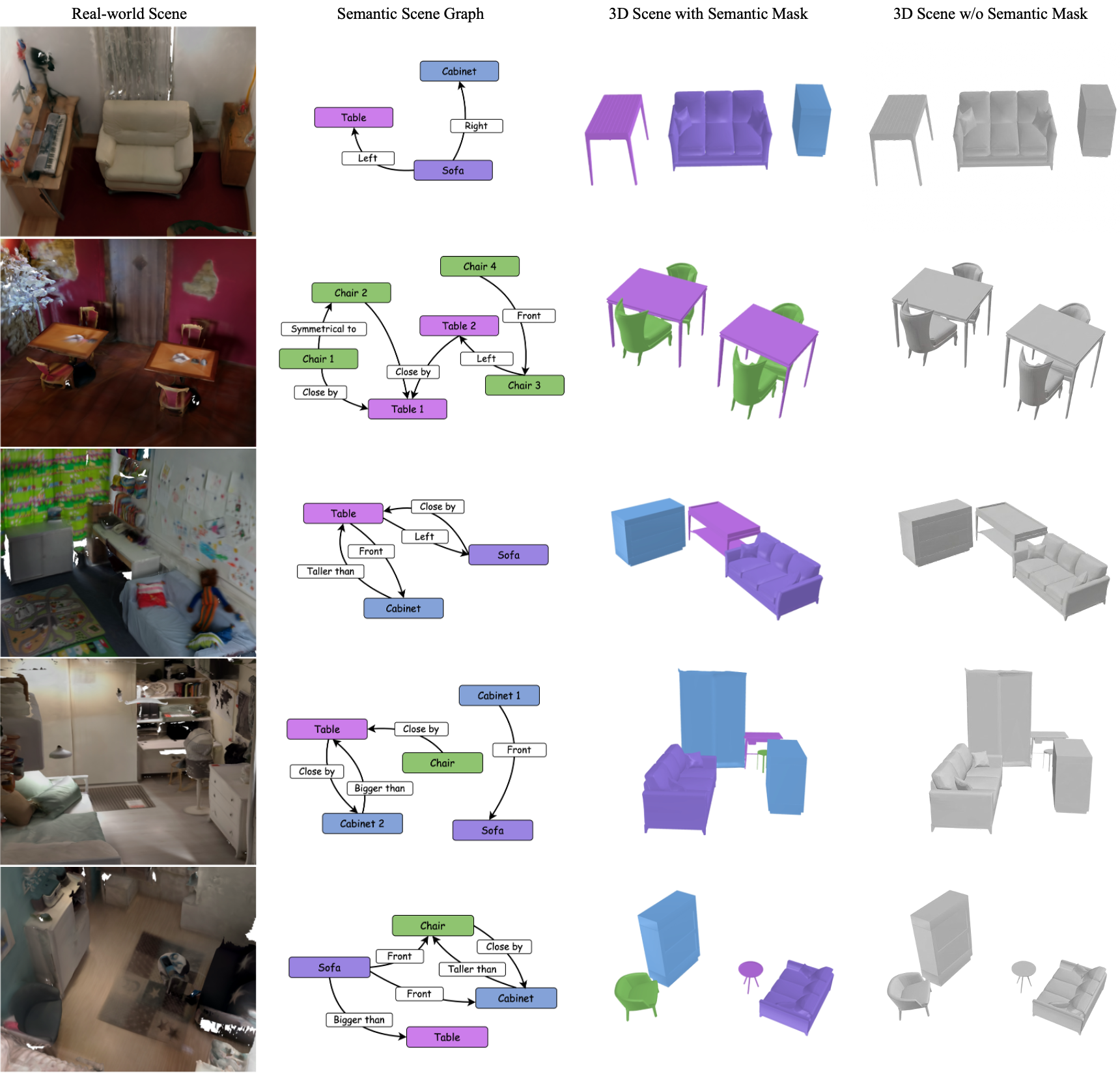}
\caption{
    Visual comparison of the real and virtual scene. To capture the entire compositional arrangement of the real-world environment, we visualize the real scene above using point clouds included in 3RScan.
}
\label{fig:qualitative3}
\end{figure*}

\cref{fig:qualitative} shows a qualitative evaluation of previous approaches that utilize text-based graph as a baseline. We compared how context-consistently the generated 3D scenes align with scene graph input. Graph-to-3D fails to maintain consistency with the scene graph, resulting in incomplete meshes. CommonScenes and EchoScene performed well in capturing object detail and variety, but struggled to fully understand the relationships within the scene. In contrast, our method generates scenes that are more context-consistent with the scene graph input than existing methods, even for challenging metrics. 
We further visually compare between 3RScan's real-world space and virtual scene in \cref{fig:qualitative3}. For this visualization, we generated a 3D scene through shape retrieval. 
In this way, users can utilize their physical surroundings as a foundation to create virtual spaces, facilitating the recording of spatial experiences.

\begin{table}[t]
\caption{
Inference runtime for each module of SceneLinker. 
}
\centering
\resizebox{0.8\columnwidth}{!}{%
\begin{tabular}{@{}cccc@{}}
\toprule
Runtime      & \begin{tabular}[c]{@{}c@{}}3D Entity \\ Estimation\end{tabular} & \begin{tabular}[c]{@{}c@{}}CCFA-based \\ GCN\end{tabular} & \begin{tabular}[c]{@{}c@{}}Graph-VAE with \\ JSL block\end{tabular} \\ \midrule
Mean {[}s{]} & 0.15s                                                           & 0.05s                                                     & 1.03s                                                               \\ \bottomrule
\end{tabular}%
}
\label{tab:runtime}
\end{table}

\subsection{Runtime}

We report the runtime of our system in \cref{tab:runtime}. During the 3D scene graph prediction process, 3D entity estimation takes 0.15s, while CCFA-based GCN computation requires only 0.05s. Additionally, generating a graph-based 3D scene takes 1.03s. In comparison, existing diffusion-based prior works such as CommonScenes and EchoScene require around 26.56s and 50.07s, respectively. Note that, MMGDreamer fully adopts EchoScene’s dual-branch diffusion model. 
Our approach demonstrates 25x faster inference speed than exist state-of-the-art method while maintaining superior generation performance.

Our work focuses on generating scenes with static instances, eliminating the need to iteratively track dynamic objects. Since the full process is performed only once through the user's initial spatial scanning, our system does not prioritize real-time processing as a primary design goal. Nevertheless, SceneLinker adopts a VAE-based network design that demonstrates faster generation speed compared to diffusion methods, showcasing significant potential for application in MR content.

\section{Discussion}
\subsection{Analysis and Implication}

Through experiments, we have identified the following technical and practical insights of SceneLinker.
First, the attention-based graph network design demonstrates robust graph prediction performance, particularly in complex and cluttered environments. Our Cross-Check Feature Attention (CCFA) network effectively refines relationships between adjacent objects by iteratively cross-checking messages between nodes and edges. Unlike organized studio environments, real-world user spaces are often cluttered or contain objects that are occluded or intertwined. As demonstrated by the experimental results in \cref{tab:sg_20} and \cref{tab:sg_160}, our model exhibits high stability in object-predicate pairwise relationship prediction, suggesting that it can robustly infer objects and relationships by considering the surrounding context even when visual information is incomplete. This indicates that when a user wearing AR/AI glasses moves through a cluttered living room or office, the system can accurately understand the global spatial structure at a concise graph level and consistently augment virtual content at the corresponding locations.

Second, we found that dominantly reflecting the bounding box in the training scheme is crucial, which contains the location of objects. As shown in \cref{tab:ablation}, the ablation results for the JSL block-based network, which fully utilizes both shape and bounding box modalities, demonstrate that layout features play a significant role in generating 3D scenes. It also proves that the strategy of CLIP-based preprocessing and integrating layouts with shapes in a single-branch is beneficial for virtualizing 3D scenes. As a result, our model, which accounts for these factors, generates layout-consistent 3D scenes with improved performance, utilizing fewer modalities compared to MMGDreamer. Consequently, this plays a key role in minimizing collision artifacts where virtual objects penetrate real walls or overlap with furniture in MR environments, thereby preventing disruption to user immersion.

Third, we analyze the practical impact of performance improvements in hard constraints on user experience. As shown in \cref{tab:generation_2}, for easy positional relationships (e.g. \texttt{left/right}, \texttt{front/behind}), most state-of-the-art models have reached saturation with high performance exceeding 0.97. However, SceneLinker recorded performance improvements of over 14\% and 7\% on hard metrics such as ``\texttt{close by}'' and ``\texttt{symmetrical}'' respectively, compared to existing SOTA methods. From the perspective of real-world MR applications, if these subtle relationships are broken, users perceive the virtual scene as unnatural space. As a result, the performance improvement demonstrated by our model in this domain suggests a step forward in enabling users to accept the virtual space as a plausible environment.

\begin{figure}
\centering
 \includegraphics[width=1.0\columnwidth]{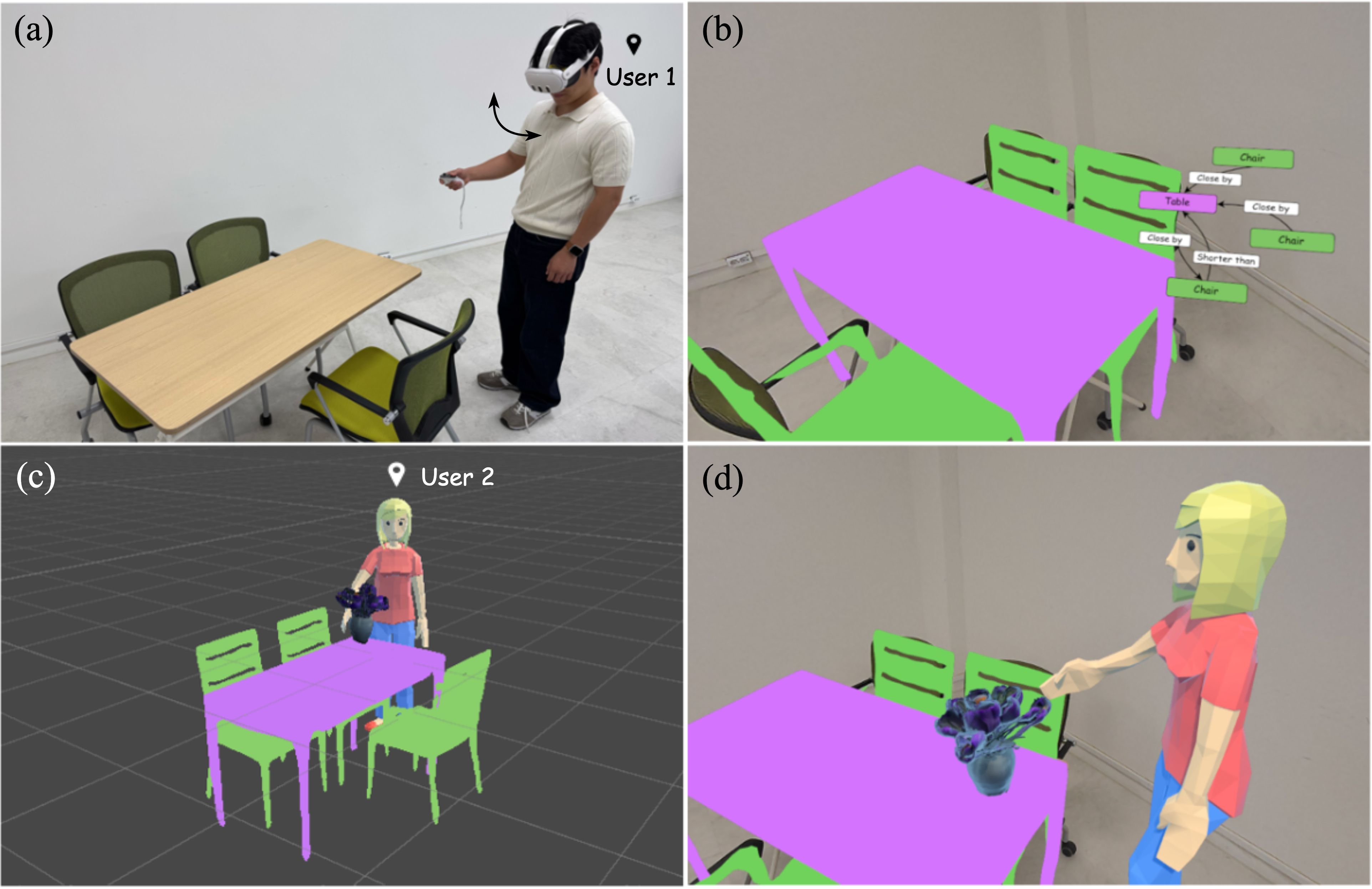}
 \caption{
SceneLinker is designed to operate in actual MR environments, showcasing its potential for deployment in AR/VR applications such as remote spatial sharing and collaborative editing. }
 \label{fig:app}
\end{figure}

\subsection{Application}
The primary advantage of SceneLinker is its ability to generate interactive 3D scenes from RGB sequences effortlessly. Unlike rendering-based methods such as 3D Gaussian Splatting \cite{3dgs}, our mesh-based representation inherently supports interactive operations including object manipulation, collision detection, and physics-aware placement, making it well-suited for immersive MR applications.
\cref{fig:app} illustrates a simple application scenario where our proposed model can be implemented using a Meta Quest 3. As depicted in \cref{fig:app}(a), when a user captures an RGB sequence through passthrough streaming, the corresponding scene graph and 3D scene are generated \cref{fig:app}(b). 
Subsequently, a remote collaborator can explore the generated scene within the VR environment and add a new object \cref{fig:app}(c). These modifications are then synchronized in real-time with the AR environment \cref{fig:app}(d), allowing the original user to continuously monitor the scene as it is updated by the collaborator. This process concisely captures entities using a simple graph structure while allowing scene updates, facilitating collaborative spatial authoring applications. It indicates that, SceneLinker can serve as a backbone for a broad range of multi-user MR applications, from remote room planning to interactive AR instruction.


\subsection{Limitation}
While this work offers substantial value as a foundational research for integrating graph-based real-virtual worlds, it still faces challenges that need to be addressed. In the 3D entity estimation as described in \cref{sec:3.1}, missing oriented bounding boxes or inaccurate poses can lead to a 3D scene that does not exactly align with reality. Such registration errors can reduce the immersion of object placement or physical interactions in MR applications.
This could be improved by using depth data or enhanced box descriptors. Also, although our system performs well both quantitatively and qualitatively, its effectiveness has not yet been validated in user experience in MR content.


The 3D scenes generated in this study are primarily composed of furniture classes included in the 3D-FRONT dataset. This limitation stems from constraints in the shape representation employed during the scene generation stage. Although the 3RScan dataset contains up to 160 object classes, the meshes within this dataset often exhibit incomplete quality and high noise levels, making them unsuitable for robust shape–layout alignment training. Therefore, to ensure stable 3D scene generation and fair benchmarking, we utilized DeepSDF-based shape codes pretrained on object classes available in the 3D-FRONT dataset. Consequently, 3RScan object classes absent in 3D-FRONT are not instantiated as actual 3D meshes during generation, leading to limited visual diversity. Furthermore, artifacts may arise during the current shape instantiation stage for complex geometries that include hollow or thin structures. Notably, this constraint does not arise from a lack of expressiveness in the scene graph prediction stage, but rather from the limitations of the current shape instantiation strategy. Future work could address this limitation by incorporating class-agnostic or open-vocabulary shape embeddings for object nodes. 
Alternatively, retrieval-based methods could be combined with generative models to produce more complex and detailed real-world scenes.

\section{Conclusion and Future Work}

In this paper, we introduce SceneLinker, which generates context-consistent 3D scenes from RGB sequences via concise graph representation. Our CCFA network achieved superior performance in propagating graph features compared to other methods. Moreover, through the graph-VAE architecture with the JSL block, we generate a virtual space that aligns precisely with the scene graph. Experiments demonstrate that our model outperforms state-of-the-art methods even in complex indoor environments and with challenging scene graph constraints. 
This highlights the effectiveness of our model in managing spatial information with scene graphs and robustly performing scene virtualization. 

For future work, we aim to enhance our framework to generate more photorealistic and interactive 3D scenes by integrating the texture post-processing module \cite{mat} to apply user-guided texture maps.
Additionally, we plan to validate our framework by applying it to AR/VR content through user experiments. We can significantly enhance immersive experiences on mobile devices or AR glasses by incorporating user-centric psychological factors into SceneLinker. Through these validations, our solution will demonstrate both academic value and strong applicability in industrial settings. By seamlessly linking the real and virtual 3D worlds at a simplified graph level, our model holds significant potential to drive advancements in spatial MR experiences.

\begin{figure}
\centering
 \includegraphics[width=1.0\columnwidth]{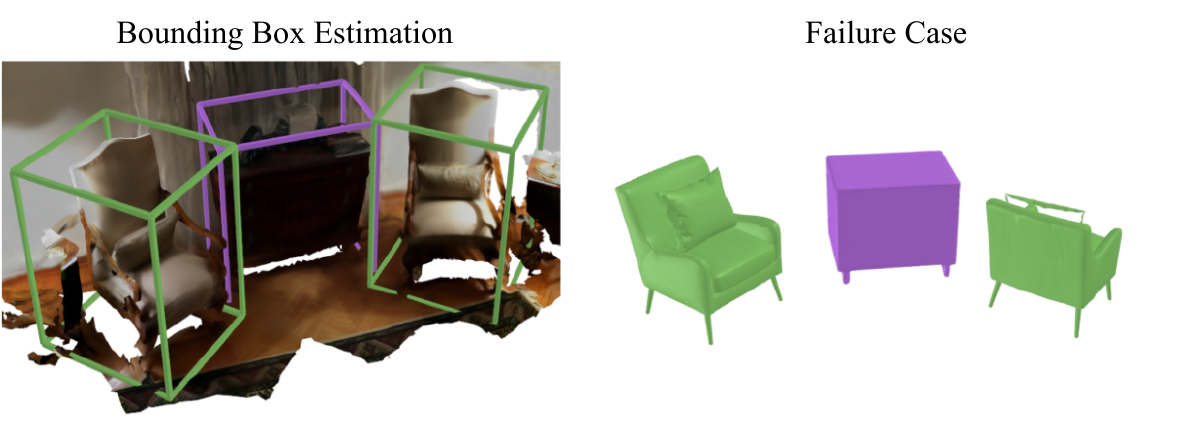}
 \caption{
Failure case. Since SceneLinker relies on 3D oriented bounding box information for 3D scene generation, it can fail if the layout feature is unstable.
 }
 \label{fig:fail_case}
\end{figure}

\acknowledgments{
We are sincerely grateful to our colleagues and the anonymous reviewers for their valuable insights and constructive feedback.
This work was supported by the IITP(Institute of Information \& Communications Technology Planning \& Evaluation)-ITRC(Information Technology Research Center) grant funded by the Korea government(Ministry of Science and ICT)(IITP-2026-RS-2024-00436398).
This work was supported by the National Research Foundation of Korea(NRF) grant funded by the Korea government(MSIT) (No. RS-2021-NR059444).
This paper was supported by Korea Institute for Advancement of Technology(KIAT) grant funded by the Korea Government(MOTIE) (RS-2025-02304167, HRD Program for Industrial Innovation). }

\bibliographystyle{abbrv-doi-hyperref}

\bibliography{template}

\appendix 







\end{document}